\documentclass[11pt,a4paper]{article}
\usepackage{bm}
\usepackage[hyperref]{naaclhlt2018}
\usepackage{times}
\usepackage{latexsym}
\usepackage{subfiles}
\usepackage{xspace}
\usepackage{booktabs}
\usepackage{graphicx}
\usepackage{enumitem}

\usepackage{url}
\usepackage{multirow}
\usepackage{makecell}

\aclfinalcopy 

\newcommand{\lcb}{{\tt {\char '173}}}   

\newcommand{\currency}{\textit{condition state}\xspace}
\newcommand{\currencynoit}{condition state\xspace}
\newcommand{\recency}{\textit{diagnosis recency}\xspace}
\newcommand{\recencynoit}{diagnosis recency\xspace}

\newcommand{\datasetname}{\textsc{RSDD}-Time\xspace}

\title{{RSDD-Time}: Temporal Annotation of Self-Reported Mental Health Diagnoses}

\author{Sean MacAvaney*, Bart Desmet$\bm{^\dagger}$*, Arman Cohan*, Luca Soldaini*,\\
  \textbf{Andrew Yates$\bm{^\ddagger}$*, Ayah Zirikly$\bm{^\S}$, Nazli Goharian*}\\
  \vspace{3mm}
  \begin{tabular}{*{2}{>{\centering}p{.5\textwidth}}}
\tabularnewline
{*}IR Lab, Georgetown University, US & $^\dagger$LT3, Ghent University, BE \tabularnewline
{\tt \lcb firstname\rcb@ir.cs.georgetown.edu} & {\tt bart.desmet@ugent.be} \tabularnewline
\tabularnewline
$^\ddagger$Max Planck Institute for Informatics, DE & $^\S$ National Institutes of Health, US \tabularnewline
{\tt ayates@mpi-inf.mpg.de} & {\tt ayah.zirikly@nih.gov}
\end{tabular}
}

\date{}

\begin{document}
\maketitle
\begin{abstract}
    Self-reported diagnosis statements have been widely employed in studying language related to mental health in social media. However, existing research has largely ignored the temporality of mental health diagnoses. In this work, we introduce \datasetname: a new dataset of 598 manually annotated self-reported depression diagnosis posts from Reddit that include temporal information about the diagnosis. Annotations include whether a mental health condition is present and how recently the diagnosis happened. Furthermore, we include exact temporal spans that relate to the date of diagnosis. This information is valuable for various computational methods to examine mental health through social media because one's mental health state is not static. We also test several baseline classification and extraction approaches, which suggest that extracting temporal information from self-reported diagnosis statements is challenging.
\end{abstract}

\section{Introduction}
\label{sec:intro}


Researchers have long sought to identify early warning signs of mental health conditions to allow for more effective treatment~\cite{feightner1990early}. Recently, social media data has been utilized as a lens to study mental health~\cite{coppersmith2017scalable}.
Data from social media users who are identified as having various mental health conditions can be analyzed to study common language patterns that indicate the condition; language use could give subtle indications of a person's wellbeing, allowing the identification of at-risk users. Once identified, users could be provided with relevant resources and support.

While social media offers a huge amount of data, acquiring manually-labeled data relevant to mental health conditions is both expensive and not scalable. However, a large amount of labeled data is crucial for classification and large-scale analysis. To alleviate this problem, NLP researchers in mental health have used unsupervised heuristics to automatically label data based on self-reported diagnosis statements such as ``I have been diagnosed with depression''~\cite{Choudhury2013PredictingDV,coppersmith2014quantifying,coppersmith2015adhd,yates2017depression}.

A binary status of a user's mental health conditions does not tell a complete story, however. People's mental condition changes over time~\cite{wilkinson2010spirit}, so the assumption that language characteristics found in a person's social media posts historically reflects their current state is invalid. For example, the social media language of an adult diagnosed with depression in early adolescence might no longer reflect any depression. Although the extraction of temporal information has been well-studied in the clinical domain~\cite{lin2016Improving,bethard2017temporal,dligach2017temporal}, temporal information extraction has remained largely unexplored in the mental health domain. Given the specific language related to self-reported diagnoses posts and the volatility of mental conditions in time, the time of diagnosis provides critical signals on examining mental health through language.

To address this shortcoming of available datasets, we introduce \datasetname: a dataset of temporally annotated self-reported diagnosis statements, based on the Reddit Self-Reported Depression Diagnosis (RSDD) dataset~\cite{yates2017depression}. \datasetname includes 598 diagnosis statements that are manually annotated to include pertinent temporal information. In particular, we identify if the conditions are current, meaning that the condition is apparently present according the the self-reported diagnosis post. Next, we identify how recently a particular diagnosis has occurred. We refer to these as \currency and \recency, respectively. Furthermore, we identify the time expressions that relate to the diagnosis, if provided.

In summary, our contributions are: \textit{(i)} We explain the necessity of temporal considerations when working with self-reported diagnoses. \textit{(ii)} We release a dataset of annotations for 598 self-reported depression diagnoses. \textit{(iii)} We provide and analyze baseline classification and extraction results.


\paragraph{Related work}
Public social media has become a lens through which mental health can be studied as it provides a public narration of user activities and behaviors \cite{conway2016social}. Understanding and identifying mental health conditions in social media (e.g., Twitter and Reddit) has been widely studied \cite{Choudhury2013PredictingDV,Coppersmith2014MeasuringPT,de2014mental,mitchell2015quantifying,gkotsis-EtAl:2016:CLPsych1,yates2017depression}. To obtain ground truth knowledge for mental health conditions, researchers have used crowdsourced surveys and heuristics such as self-disclosure of a diagnosis \cite{Choudhury2013PredictingDV,Tsugawa2015RecognizingDF}.
The latter approach uses high-precision patterns such as ``I was diagnosed with depression.'' Only statements claiming an actual diagnosis are considered because people sometimes use phrases such as ``I am depressed'' casually. In these works, individuals self-reporting a depression diagnoses are presumed to be depressed.
Although the automated approaches have yielded far more users with depression than user surveys (tens of thousands, rather than hundreds), there is no indication of whether or not the diagnosis was recent, or if the conditions are still present. In this work, we address this by presenting manual annotations of nearly 600 self-reported diagnosis posts. This dataset is valuable because it allows researchers to train and test systems that automatically determine \recencynoit and \currencynoit information.

\section{Data}
\label{sec:data}
For the study of temporal aspects of self-reported diagnoses, we develop an annotation scheme\footnote{Available at \url{http://ir.cs.georgetown.edu/resources/}}
and apply it to a set of 598 diagnosis posts randomly sampled from the Reddit Self-Reported Depression Diagnosis (RSDD) dataset \citep{yates2017depression}.
In the annotation environment, the diagnosis match is presented with a context of 300 characters on either side. A window of 150 characters on either side was too narrow, and having the whole post as context made annotation too slow, and rarely provided additional information.

\paragraph{Annotation scheme}
Two kinds of text spans are annotated: diagnoses (e.g., ``I was diagnosed'') and time expressions that are relevant to the diagnosis (e.g., ``two years ago''). On diagnosis spans, the following attributes are marked:

\begin{itemize}[itemindent=8pt,leftmargin=0pt]
	\setlength\itemsep{0pt}
	\item \textbf{Diagnosis recency} determines when the diagnosis occurred (not the onset of the condition). Six categorical labels are used: very recently (up to 2 months ago), more than 2 months but up to 1 year ago, more than 1 year but up to 3 years ago, more than 3 years ago, \textit{unspecified} (when there is no indication), and \textit{unspecified but not recent} (when the context indicates that the diagnosis happened in the past, yet there is insufficient information to assign it to the first four labels).
	\item For \textbf{\currencynoit}, the annotator assesses the context for indications of whether the diagnosed condition is still current or past. The latter includes cases where it is reported to be fully under control through medication. We use a five-point scale (\textit{current, probably current, unknown, probably past} and \textit{past}). This can be mapped to a three-point scale for coarse-grained prediction (i.e. moving \emph{probable} categories to the center or the extremes).
	\item When a diagnosis is presented as uncertain or incorrect, we mark it as \textbf{diagnosis in doubt}. This can be because the diagnosis is put into question by the poster (e.g., ``I was diagnosed with depression before they changed it to ADHD''), or it was later revised.
	\item Occasionally, incorrect diagnosis matches are found in RSDD. These are marked as \textbf{false positive}. This includes diagnoses for conditions other than depression or self-diagnosis that occur in block quotes from other posts. False positive posts are not included in the analyses below.
\end{itemize}

Time expressions indicating the time of diagnosis are marked similarly to the TIMEX3 specification~\cite{pustejovsky2005specification}, with the additional support for ages, years in school, and references to other temporal anchors. Because of these additions, we also annotate prepositions pertaining to the temporal expression when present (e.g., `at 14', `in 2004'). Each span also has an indication of how their associated diagnosis can be assigned to one of the \recency labels.
\textbf{Explicit} time expressions allow immediate assignment given the post date (e.g., yesterday, last August, in 2006).
If the recency can be inferred assuming a poster's age at post time is known, it is \textbf{inferable from age} (e.g., at 17, in high school).
A poster's age could be established using mentions by the author, or estimated with automatic age prediction.


\paragraph{Inter-annotator agreement}
After an initial annotation round with 4 annotators that allowed for the scheme and guidelines to be improved, the entire dataset was annotated by 6 total annotators with each post being at least double annotated; disagreements were resolved by a third annotator where necessary. We report pairwise inter-annotator agreement in Table~\ref{tab:iaa}. Cohen's kappa is linearly weighted for ordinal categories (\currency and \recency).

\begin{table}
\small\centering
\begin{tabular}{@{}llrr@{}}
\toprule
\bf Span & \bf Attribute & \% & $\kappa$ \\
\midrule
\multirow{4}{*}{diagnosis} & false positive & 0.97 & 0.43 \\
 & diagnosis in doubt & 0.97 & 0.22 \\
 & \currencynoit & 0.52 & 0.41 \\
 & \recencynoit & 0.66 & 0.64 \\
\midrule
\multirow{2}{*}{time} & explicit & 0.91 & 0.81 \\
 & inferable from age & 0.93 & 0.82 \\
\bottomrule
\end{tabular}
\caption{Inter-annotator agreement by average pairwise agreement (\%) and weighted Cohen's kappa ($\kappa$).}
\label{tab:iaa}
\end{table}

Agreement on false positives and doubtful diagnoses is low.
For future analyses that focus on detecting potential misdiagnoses, further study would be required to improve agreement, but it is tangential to the focus on temporal analysis in this study.

Estimating the state of a condition is inherently ambiguous, but agreement is moderate at $0.41$ weighted kappa. The five-point scale can be backed off to a three-point scale, e.g. by collapsing the three middle categories into \emph{don't know}. Pairwise percent agreement then improves from $0.52$ to $0.68$. The recency of a diagnosis can be established with substantial agreement ($\kappa = 0.64$). Time expression attributes can be annotated with almost perfect agreement.

\paragraph{Availability}
The annotation data and annotation guidelines are available at \url{https://github.com/Georgetown-IR-Lab/RSDD-Time}. The raw post text is available from the RSDD dataset via a data usage agreement (details available at \url{http://ir.cs.georgetown.edu/resources/rsdd.html}).

\section{Corpus analysis}
\label{sec:analysis}

{
\renewcommand{\arraystretch}{1.3}
\begin{table}
\setlength{\tabcolsep}{2pt}
\small
\begin{tabular}{@{}ll@{}}
\toprule
\bf Attribute & \bf Count \\
\midrule
false positive & 25 out of 598 \\
diagnosis in doubt & 16 out of remaining 573 \\ [2pt]
\currencynoit & \makecell[l]{current (254), prob. current (64), \\ unknown (225), prob. past (29), past (26)} \\[8pt]
\recencynoit & \makecell[l]{unspec. (232), unspec. but past (176), \\ recent (27), $>$2m-1y (37), \\ $>$1y-3y (29), $>$3y (97)} \\ [14pt]
time expression & \makecell[l]{explicit (144), inferable from age (101), \\ non-inferable (47), n/a (306)} \\
\bottomrule
\end{tabular}
\caption{Attribute counts in the \datasetname dataset.}
\label{tab:stats}
\end{table}
}

\begin{figure}
  \centering
\includegraphics[width=0.8\linewidth]{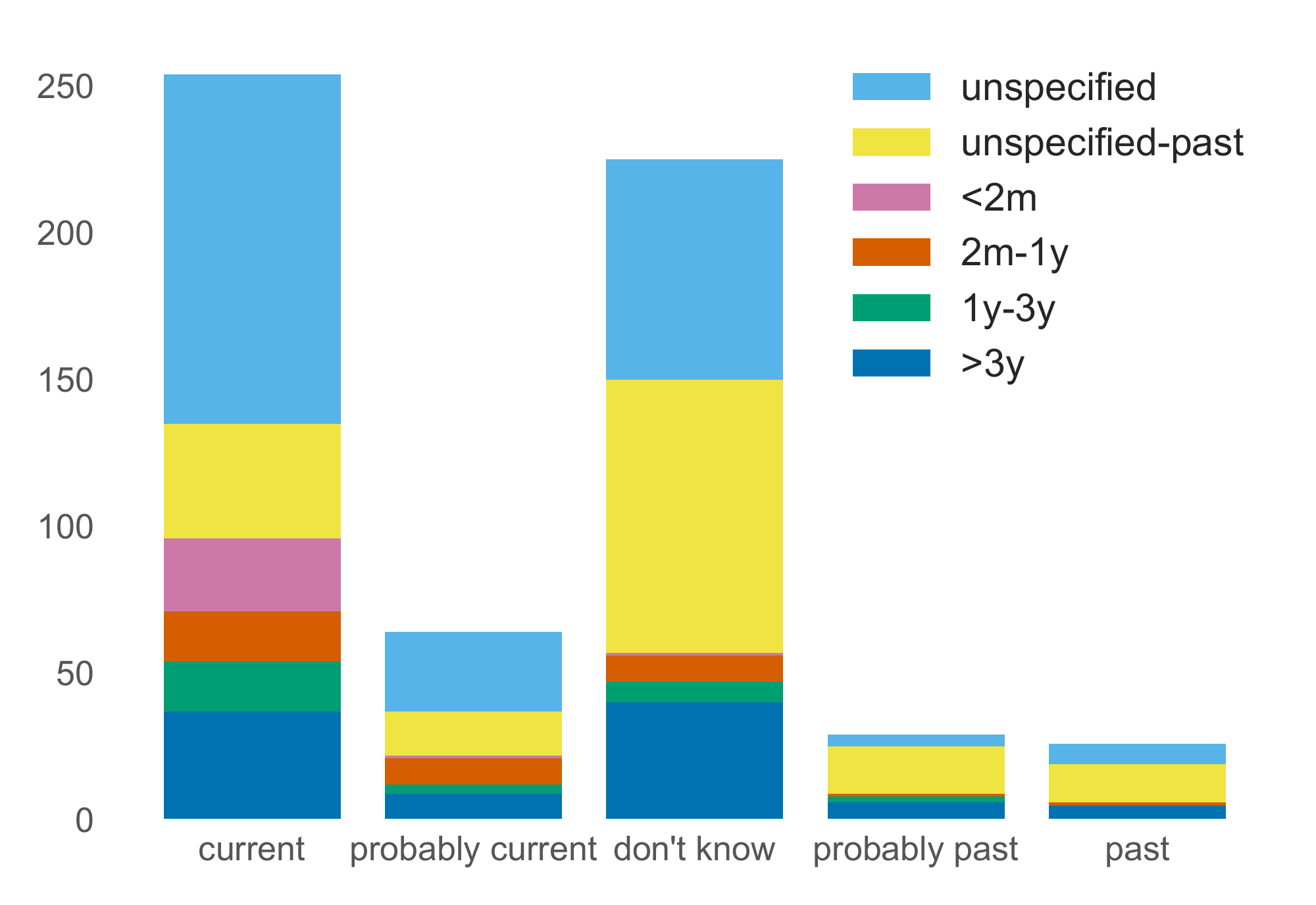}
\caption{Incidence and interaction of \currency (columns) and \recency (colors).}
\label{fig:stats}
\end{figure}

Counts for each attribute are presented in Table~\ref{tab:stats}.
Figure \ref{fig:stats} shows the incidence and interaction between \currency and \recency in our dataset. About half the cases have a \currency that is current, but interestingly, there are also many cases (55) where the diagnosis relates (at least probably) to the past. There is also a large number of cases (225) where it is not clear from the post whether the condition is current or not. This further shows that many self-reported diagnosis statements may not be current, which could make a dataset noisy, depending on the objective. For \recency, we observe that the majority of diagnosis times are either unspecified or happened in the unspecified past. For 245 cases, however, the \recency can be inferred from the post, usually because there is an explicit time expression (59\% of cases), or by inferencing from age (41\%).
Next, we investigate the interaction between \currency and \recency. We particularly observe that the majority of past conditions (rightmost two columns) are also associated with a \recency of more than 3 years ago or of an unspecified past. On the other hand, many current conditions (leftmost column) have an unspecified diagnosis time. This is expected because individuals who specifically indicate that their condition is not current also tend to specify when they have been first diagnosed, whereas individuals with current conditions may not mention their time of diagnosis.

\section{Experiments}
\label{sec:method}

To gain a better understanding of the data and provide baselines for future work to automatically perform this annotation, we explore methods for attribute classification for \recency and \currency, and rule-based diagnosis time extraction. We split the data into a training dataset (399 posts) and a testing dataset (199 posts). We make this train/test split available for future work in the data release. For our experiments, we then disregard posts that are labeled as \textit{false positive} (yielding 385 posts for training and 188 for testing), and we only consider text in the context window with which the annotator was presented.

\subsection{Diagnosis recency and \currencynoit classification}
\label{sec:meth-diag}

We train several models to classify \recency and \currency. In each we use basic bag-of-character-ngrams features. Character ngrams of length 2-5 (inclusive) are considered, and weighted using \textit{tf-idf}. For labels, we use the combined classes described in Section~\ref{sec:data}. To account for class imbalance, samples are weighed by the inverse frequency of their category in the training set.

We compare three models: logistic regression, a linear-kernel Support Vector Machine (SVM), and Gradient-Boosted ensemble Trees (GBT)~\cite{chen2016xgboost}. The logistic regression and SVM models are $\ell_2$ normalized, and the GBT models are trained with a maximum tree depth of 3 to avoid overfitting.

We present results in Table~\ref{tab:model-results}. The GBT method performs best for \recency classification, and logistic regression performs best for \currency classification.
This difference could be due to differences in performance because of skew. The \currency data is more skewed, with \textit{current} and \textit{don't know} accounting for almost 80\% of the labels.

\begin{table}
\small\centering
\begin{tabular}{@{}lrrrrrr@{}}
\toprule
&\multicolumn{3}{c}{\bf Diagnosis Recency}&\multicolumn{3}{c}{\bf Condition State} \\
\cmidrule(lr){2-4}\cmidrule(lr){5-7}
&\bf P &\bf R &\bf F1 &\bf P &\bf R &\bf F1 \\
\midrule
Logistic Reg.       &   0.47 &   0.35 &   0.37 &   0.45 &\bf0.45 &\bf0.44 \\
Linear SVM          &   0.23 &   0.23 &   0.21 &\bf0.68 &   0.40 &   0.40 \\
GBT                 &\bf0.56 &\bf0.42 &\bf0.46 &   0.35 &   0.38 &   0.36 \\
\bottomrule
\end{tabular}
\caption{Macro-averaged classification results for \recency and \currency using \textit{tf-idf} vectorized features for various baseline models.}
\label{tab:model-results}
\vspace{-1em}
\end{table}

\subsection{Time expression classification}

To automatically extract time expressions, we use the rule-based SUTime library~\cite{chang2012sutime}. Because diagnoses often include an age or year in school rather than an absolute time, we added rules specifically to capture these time expressions. The rules were manually generated by examining the training data, and will be released alongside the annotations.

\datasetname temporal expression annotations are only concerned with time expressions that relate to the diagnosis, whereas SUTime extracts all temporal expressions in a given text. We use a simple heuristic to resolve this issue: simply choose the time expression closest to the post's diagnosis by character distance. In the case of a tie, the heuristic arbitrarily selects the leftmost expression. This heuristic will improve precision by eliminating many unnecessary temporal expressions, but has the potential to reduce precision by eliminating some correct expressions that are not the closest to the diagnosis.

Results for temporal extraction are given in Table~\ref{tab:sutime}. Notice that custom age rules greatly improve the recall of the system. The experiment also shows that the \textit{closest} heuristic improves precision at the expense of recall (both with and without the age rules). Overall, the best results in terms of F1 score are achieved using both the \textit{closest} heuristic and the age rules. A more sophisticated algorithm could be developed to increase the candidate expression set (to improve recall), and better predict which temporal expressions likely correspond to the diagnosis (to improve precision).

\begin{table}
\small\centering
\begin{tabular}{@{}lrrr@{}}
\toprule
&\bf P &\bf R &\bf F1 \\
\midrule
SUTime                          &   0.17 &   0.59 &   0.26 \\
+ age rules                     &   0.20 &\bf0.81 &   0.32 \\
+ closest heuristic             &   0.33 &   0.51 &   0.40 \\
+ closest heuristic + age rules &\bf0.44 &   0.69 &\bf0.53 \\
\bottomrule
\end{tabular}
\caption{Results using SUTime, with additional rules for predicting age expressions and when limiting the candidate expression set using the \textit{closest} heuristic.}
\label{tab:sutime}
\vspace{-1em}
\end{table}

\section{Conclusion}
In this paper, we explained the importance of temporal considerations when working with language related to mental health conditions. We introduced \datasetname, a novel dataset of manually annotated self-reported depression diagnosis posts from Reddit. Our dataset includes extensive temporal information about the diagnosis, including when the diagnosis occurred, whether the condition is still current, and exact temporal spans. Using \datasetname, we applied rule-based and machine learning methods to automatically extract these temporal cues and predict temporal aspects of a diagnosis. While encouraging, the experiments and dataset allow much room for further exploration.

\bibliography{references}
\bibliographystyle{acl_natbib}

\end{document}